\documentclass[runningheads]{llncs}

\usepackage[T1]{fontenc}
\usepackage{graphicx}
\usepackage{amsmath,amssymb}
\usepackage{booktabs}
\usepackage{algorithm}
\usepackage{algorithmic}
\usepackage{hyperref}
\usepackage{multirow}
\usepackage[table]{xcolor}

\begin{document}

\title{A Differentiable Optimization Framework for Registering Sequential Bounding Boxes with Point Cloud Stream}
% \titlerunning{Differentiable Box--Point Cloud Registration}

\author{Xuesong Li\inst{1}, Jinguang Tong\inst{1}, Jie Hong\inst{2}}
\institute{CSIRO, The University of Hong Kong}

\maketitle

\begin{abstract}
Refining a sequence of coarse 3D bounding boxes against a LiDAR point-cloud
stream demands tracks that are geometrically accurate (high IoU) and
temporally coherent (low roughness), preferably without training data. The
usual recipe keeps the two concerns apart: register each frame independently,
then smooth the trajectory afterwards with a Kalman~RTS or Savitzky--Golay
filter. Smoothing displaces boxes from a geometric optimum and never
re-optimises, so it trades accuracy for smoothness. We instead fold the
temporal smoothness constraint into a training-free registration objective
and solve for all poses jointly with L-BFGS. The payoff depends on how well
the object is seen. On well-observed tracks it is large: within the
low-roughness budget, the joint objective beats both post-hoc smoothers on
paired multi-seed statistics and cuts roughness several-fold relative to
frame-wise registration at matched accuracy. Treating visibility as an
experimental variable exposes the limit. The advantage decays monotonically
as views become one-sided, until it is indistinguishable from zero for
near-edge-on objects and slightly negative under a ray-cast simulator with
range-dependent density and ego motion, where the decoupled pipeline is in
fact ahead at tight roughness budgets. We locate that boundary and trace it
to one term: orientation alignment ties yaw to the estimated velocity and
fails once that estimate is noisy. A ground-truth-free rule can choose the
temporal scale and keep every track inside the roughness budget.
\end{abstract}


\section{Introduction}\label{sec:intro}

Accurate 3D bounding box annotations are essential for training perception systems in autonomous driving, yet manual labeling of LiDAR point clouds is costly and time-consuming~\cite{sun2020waymo,caesar2020nuscenes,li2019three}. Auto-labeling pipelines address this by automatically refining noisy bounding box proposals generated by onboard detectors~\cite{qi2021offboard,yang2021auto4d,10056417,li2020real}. A critical step in these pipelines is \emph{registration}: aligning each bounding box to its corresponding point cloud observations across a temporal sequence. Existing approaches to this registration problem fall into two categories. Learning-based methods~\cite{yang2021auto4d,fan2023ctrl,li2019detection,fusion4dal2025,hong2026pointcam} train neural networks to predict refined box poses, achieving high accuracy but requiring large labeled datasets and expensive retraining for new domains. Classical registration methods such as ICP~\cite{besl1992icp} align point sets geometrically but treat each frame independently, ignoring the temporal structure of sequential observations and the geometric properties of bounding boxes.

We propose a \emph{training-free differentiable optimization framework} that directly models all geometric and temporal constraints in a unified objective function. Our framework jointly optimizes all bounding box poses in a sequence by combining four differentiable loss terms: (1)~a \emph{closeness} term that minimizes point-to-visible-face distances, (2)~an \emph{enclosure} term that penalizes points outside their boxes using L1 distance, (3)~a \emph{smoothness} term that penalizes trajectory acceleration, and (4)~an \emph{alignment} term that enforces consistency between box orientation and movement direction. The complete objective is optimized via L-BFGS on a CPU, with no training. By default the alignment term is disabled (Sec.~\ref{sec:ablation}), giving a three-term objective.

Temporal constraints should shape the geometric fit rather than be applied after it. That is the argument for solving jointly. What joint optimisation delivers is not a raw accuracy gain over the noisy initialisation, since frame-wise registration alone already recovers most of that, but a better \emph{accuracy--smoothness operating point}, which we quantify with multi-seed paired statistics. The harder question is when that operating point is worth having: under what observation conditions does the joint design pay?

That question decides the paper. Where the sensor sees most of the object,
joint optimisation beats post-hoc smoothing by a wide margin. Swap in a
ray-cast simulator with one-sided visibility, range-dependent density and ego
motion, closer to what a real LiDAR delivers, and the same comparison is a
tie; at tight roughness budgets the decoupled pipeline is ahead. Our contributions are:
\begin{itemize}
    \item A training-free differentiable objective that folds temporal
    smoothness into the geometric registration terms (closeness, enclosure)
    and optimises all box poses in a sequence jointly with L-BFGS. No labels,
    no training.
    \item An operating-point result for well-observed objects. Within the
    low-roughness budget the joint objective is more accurate than both
    Savitzky--Golay and Kalman post-hoc smoothing ($+0.032$ and $+0.059$ IoU,
    $p{<}10^{-4}$, ahead on every track), and reduces trajectory roughness
    $5.8\times$ relative to frame-wise registration at statistically
    indistinguishable IoU.
    % \item A ground-truth-free rule that sets the temporal scale per sequence from the observed frame-wise roughness, keeping $100\%$ of tracks inside the roughness budget where a fixed setting leaves up to $17\%$ outside. The same idea does not extend to selecting \emph{which} constraints to include: no cue we measured predicts the alignment term's utility, and we report that negative result.
    % \item A metric correction. The fast grid IoU estimator compresses effect
    sizes by roughly $2.5\times$ and reverses the sign of one ablation row, so
    all results here use an unbiased Monte-Carlo estimator.
\end{itemize}
\section{Related Work}\label{sec:related}

\subsection{Learning-Based Auto-Labeling}

Modern auto-labeling pipelines follow a detect--track--refine paradigm. Qi et al.~\cite{qi2021offboard} proposed offboard 3D detection from point cloud sequences, aggregating multi-frame information for higher accuracy. Auto4D~\cite{yang2021auto4d} introduced object-centric 4D auto-labeling from sequential point clouds. CTRL~\cite{fan2023ctrl} demonstrated that track-centric bidirectional refinement can surpass human annotation accuracy on the Waymo Open Dataset. More recently, Fusion4DAL~\cite{fusion4dal2025} fused LiDAR and camera modalities for offline detection, OpenBox~\cite{openbox2025} leveraged 2D vision foundation models for zero-shot 3D annotation, and ZOPP~\cite{zopp2024} proposed zero-shot offboard panoptic perception. All require training data or foundation-model inference. Our framework is complementary: it provides a training-free refinement module that can be integrated into any pipeline's box refinement stage.

\subsection{Optimization-Based Registration}

The Iterative Closest Point (ICP) algorithm~\cite{besl1992icp} and its variants~\cite{rusinkiewicz2001icp,segal2009gicp} are the classical approach to point cloud registration. Go-ICP~\cite{yang2016goicp} guarantees global optimality via branch-and-bound search over SE(3) but is computationally expensive. ICP-Flow~\cite{lin2024icpflow} recently demonstrated that ICP with histogram-based initialization achieves state-of-the-art learning-free scene flow estimation on the Waymo dataset. DC-Reg~\cite{dcreg2026} proposed globally optimal registration via difference of convex programming. These methods align point sets to point sets; our problem is to align \emph{bounding boxes} to point clouds---a formulation that can exploit box geometry (faces, enclosure, known dimensions) and temporal consistency across a sequence.

\subsection{Differentiable Optimization in 3D Perception}

Differentiable optimization has been increasingly applied to 3D perception tasks. Deep Closest Point~\cite{wang2019dcp} introduced learned attention with a differentiable SVD for registration. EulerFlow~\cite{vedder2025eulerflow} models scene flow as a continuous ODE optimized via neural priors, achieving state-of-the-art self-supervised performance. Diff$^2$I2P~\cite{diff2i2p2025} makes image-to-point-cloud registration fully differentiable using diffusion priors. We share the differentiable-optimization philosophy but target a different problem: registering \emph{boxes} (not point sets or images) to point cloud streams under explicit temporal constraints, with a hand-crafted geometric objective and no training.

\section{Method}\label{sec:method}

\subsection{Problem Formulation}

Given a sequence of $N$ observed point clouds $\{P_1, \ldots, P_N\}$ and corresponding initial bounding box poses $\{B_1, \ldots, B_N\}$, where each box $B_i$ is parameterized by its center position $(x_i, y_i, z_i)$ and Euler angles $(\alpha_i, \beta_i, \gamma_i)$, we seek refined poses that best align each box with its point cloud observations while maintaining temporal coherence.

The box dimensions $(l, w, h)$ are assumed known and fixed across the sequence (the same physical object observed at different times). The optimized variables are the 6-DOF poses: $\mathbf{q}_i = [x_i, y_i, z_i, \alpha_i, \beta_i, \gamma_i]$ for $i = 1, \ldots, N$. For 2D BEV registration, only $(x_i, y_i, \gamma_i)$ are optimized.

\subsection{Objective Function}

We define a total objective $\mathcal{L}_T$ as a weighted sum of four differentiable terms:
\begin{equation}\label{eq:total}
    \mathcal{L}_T = \delta \cdot \mathcal{L}_c + \omega \cdot \mathcal{L}_e + \varepsilon \cdot \mathcal{L}_s + \theta \cdot \mathcal{L}_a
\end{equation}
where $\delta, \omega, \varepsilon, \theta$ are non-negative weights. Each term is described below.

\subsubsection{Closeness Term $\mathcal{L}_c$.}

For each box $B_i$ and its point cloud $P_i$, we transform $P_i$ into the box's local coordinate frame. We then identify the \emph{visible face} on each axis by comparing the mass center of the transformed points to the box center. For each axis $d \in \{x, y, z\}$, the $K$ closest points to the visible face are selected, and their squared point-to-plane distances are averaged:
\begin{equation}\label{eq:closeness}
    \mathcal{L}_c = \frac{1}{N} \sum_{i=1}^{N} \frac{1}{3} \sum_{d=1}^{3} \frac{1}{K} \sum_{k=1}^{K} (p_{i,d,k} - f_{i,d})^2
\end{equation}
where $p_{i,d,k}$ is the $k$-th closest point coordinate along axis $d$ and $f_{i,d}$ is the visible face position. This term pulls points toward their nearest box faces.

\subsubsection{Enclosure Term $\mathcal{L}_e$.}

All observed points should lie inside their corresponding bounding box. Using L1 distance to each face:
\begin{equation}\label{eq:enclosure}
    \mathcal{L}_e = \frac{1}{N} \sum_{i=1}^{N} \frac{1}{|P_i|} \sum_{\mathbf{p} \in P_i} \sum_{d=1}^{3} \max\!\left(0,\; |p^{(d)}| - \tfrac{s_d}{2}\right)
\end{equation}
where $p^{(d)}$ is the $d$-th coordinate in the local frame and $s_d$ is the box's full extent along axis $d$ (so $s_d/2$ is the face offset). We deliberately use an L1 (rather than L2) formulation: it yields zero gradient for points already inside the box, so well-enclosed points exert no spurious pull, whereas an L2 penalty would be minimized only when points sit at the box \emph{center}---the opposite of the intended behavior.

\subsubsection{Smoothness Term $\mathcal{L}_s$.}

Temporal consistency is enforced by penalizing acceleration (second-order finite differences) in the pose trajectory:
\begin{equation}\label{eq:smoothness}
    \mathcal{L}_s = \frac{1}{N{-}2} \sum_{i=2}^{N-1} \|\Delta \mathbf{q}_i - \Delta \mathbf{q}_{i-1}\|^2
\end{equation}
where $\Delta \mathbf{q}_i = \mathbf{q}_{i+1} - \mathbf{q}_i$. This penalizes abrupt changes in velocity across all pose variables.

\subsubsection{Alignment Term $\mathcal{L}_a$.}

The box's forward direction should align with its movement direction:
\begin{equation}\label{eq:alignment}
    \mathcal{L}_a = \frac{1}{N{-}1} \sum_{i=1}^{N-1} \|\mathbf{o}_i - \mathbf{u}_i\|^2
\end{equation}
where $\mathbf{o}_i = [\cos\beta_i\cos\gamma_i,\; \cos\beta_i\sin\gamma_i,\; -\sin\beta_i]^\top$ is the orientation unit vector and $\mathbf{u}_i$ is the normalised displacement between consecutive positions. The term couples orientation to the \emph{estimated} velocity, so its usefulness follows the quality of that estimate rather than the amount of visible geometry. It contributes little when the object is well observed. It helps when views are sparse but ranges are clean. Once range noise and ego motion corrupt the velocity estimate it hurts (Sec.~\ref{sec:ablation}), so we disable it by default ($\theta{=}0$) and use the remaining three terms.

\subsection{Optimization}

The complete objective~\eqref{eq:total} is implemented in PyTorch, enabling automatic differentiation. We use L-BFGS~\cite{nocedal2006optimization} with strong Wolfe line search as the primary optimizer. Convergence is declared when $|\mathcal{L}_T^{(k)} - \mathcal{L}_T^{(k-1)}| < 10^{-6}$ or after 500 iterations. The algorithm is summarized in Algorithm~\ref{alg:optimize}.

\begin{algorithm}[t]
\caption{Differentiable Box--Point Cloud Registration}
\label{alg:optimize}
\begin{algorithmic}[1]
\REQUIRE Point clouds $\{P_i\}_{i=1}^N$, initial poses $\{\mathbf{q}_i^{(0)}\}$, box dimensions $(l,w,h)$, weights $(\delta,\omega,\varepsilon,\theta)$
\ENSURE Refined poses $\{\mathbf{q}_i^*\}$
\STATE Initialize L-BFGS optimizer with parameters $\{\mathbf{q}_i^{(0)}\}$
\FOR{$k = 1, \ldots, K_{\max}$}
    \STATE Compute $\mathcal{L}_c$, $\mathcal{L}_e$, $\mathcal{L}_s$, $\mathcal{L}_a$ via Eqs.~\eqref{eq:closeness}--\eqref{eq:alignment}
    \STATE $\mathcal{L}_T \leftarrow \delta \mathcal{L}_c + \omega \mathcal{L}_e + \varepsilon \mathcal{L}_s + \theta \mathcal{L}_a$
    \STATE Compute $\nabla_{\mathbf{q}} \mathcal{L}_T$ via autograd
    \STATE Update $\{\mathbf{q}_i\}$ via L-BFGS step
    \IF{$|\mathcal{L}_T^{(k)} - \mathcal{L}_T^{(k-1)}| < 10^{-6}$}
        \STATE \textbf{break} \COMMENT{Converged}
    \ENDIF
\ENDFOR
\RETURN $\{\mathbf{q}_i^*\}$
\end{algorithmic}
\end{algorithm}

\subsection{Weight Selection}

The enclosure weight $\omega$ has the strongest impact on accuracy, while the smoothness weight $\varepsilon$ trades IoU for smoothness along a frontier (Sec.~\ref{sec:richness}). Our default objective is the three-term form ($\theta{=}0$, i.e.\ closeness, enclosure, smoothness): the alignment term adds only $+0.002$ IoU ($p{=}0.17$) on idealized data yet costs up to $0.09$ IoU under ray-cast observation (Sec.~\ref{sec:ablation}), so dropping it is both simpler and more robust. We fix $\delta{=}2$, $\omega{=}5$ and select $\varepsilon$ by the desired roughness budget; $\varepsilon{=}0.5$ is the recommended operating point, placing the track inside the low-roughness regime ($\bar{R}\approx0.08$) at near-peak IoU. Sec.~\ref{sec:selection} gives a rule that sets $\varepsilon$ automatically.

\section{Experiments}\label{sec:experiments}

\subsection{Setup}\label{sec:setup}

\textbf{Simulation.} Known ground-truth poses let accuracy be measured exactly. A dense surface cloud (2000 points) is sampled on the box faces, placed at the ground-truth pose, and observed from a simulated ego viewpoint with visibility culling, giving 200--300 points per frame; initial boxes are perturbed with Gaussian noise ($\sigma_{\mathrm{pos}}{=}0.3$\,m, $\sigma_{\mathrm{ang}}{=}0.15$\,rad). Results are pooled over five seeds with paired $95\%$ confidence intervals.

\noindent \textbf{Observation regimes.}\label{par:regimes} One parameter controls how much of the object is seen: a point is kept when its face normal satisfies $\mathbf{n}^\top\mathbf{v} > c$ for view direction $\mathbf{v}$. At $c{=}-0.1$ even grazing faces are kept ($4.1$ faces per frame on average); $c{=}0.8$ approaches a single-face view ($1.2$ faces). We also use a higher-fidelity \emph{ray-cast} simulator with HDL-64E-like beams, self-occlusion, range-dependent density, and a moving ego sensor whose pose lifts observations to a world frame. This axis is the subject of Sec.~\ref{sec:richness}: the idealised setting is the most favourable one for our method.

\noindent \textbf{Shapes and motion.} Four categories (SUV, sedan, pedestrian, cyclist; $4.7{\times}2.0{\times}1.7$\,m down to $0.5{\times}0.5{\times}1.7$\,m) and three motion types with $N{=}30$ frames: (A)~mountain road with coupled yaw, roll and pitch; (B)~highway lane change; (C)~urban intersection with a $90^\circ$ turn and stop-and-go velocity.

\noindent \textbf{Metrics.} Accuracy is 3D volumetric IoU, estimated by uniform Monte-Carlo sampling inside one box and measuring the fraction inside the other. A faster regular-grid estimator is rank-preserving (Spearman $0.997$ against this reference) but \emph{compresses} differences, since under rotation its boundary exclusion biases identical boxes down to $\approx0.89$: it shrinks effect sizes $\approx2.5\times$ and reverses one ablation row's sign (Sec.~\ref{sec:ablation}). Every number below is therefore the unbiased Monte-Carlo IoU unless labelled \emph{grid}. Smoothness is positional trajectory roughness $\bar{R}=\frac{1}{N-3}\sum\|\Delta^3\mathbf{x}\|$ in m\,frame$^{-3}$ (the third difference of box centers, \emph{jerk} in the ride-comfort literature); $\bar{R}\le0.1$ is the \emph{low-roughness budget}, and ground-truth tracks have $\bar{R}=0.008$.

\noindent \textbf{Baselines.} (1)~Initial boxes; (2)~point-to-plane box-surface ICP, per frame, with no temporal coupling; (3)~frame-wise optimization, i.e.\ our objective with $\varepsilon{=}0$; (4)~frame-wise Savitzky--Golay smoothing; (5)~frame-wise plus a Kalman RTS smoother. Baselines (3)--(5) form a \emph{same-terms isolation}: frame-wise is our objective without the temporal term, decoupled is frame-wise plus a smoother, and joint is ours; the only variable changed is \emph{where} the temporal constraint applies.

\noindent \textbf{Operating point.} The default objective is three-term (closeness, enclosure, smoothness) with the geometric base $\delta{=}2,\,\omega{=}5$; the smoothness term is scaled by $\varepsilon$, and $\varepsilon{=}0.5$ places the track inside the budget at near-peak accuracy.

\subsection{Operating-Point Comparison}\label{sec:main}

\begin{table}[t]
\centering
\caption{Accuracy--smoothness operating point on the idealised multi-face generator, pooled over motions A,B,C and five seeds ($n{=}15$). Among methods inside the low-roughness budget (shaded), ours is the most accurate. $\pm$ denotes a $95\%$ CI half-width. Ground-truth roughness: $0.008$.}\label{tab:baselines}
\begin{tabular}{lcc}
\toprule
Method & Mean IoU $\uparrow$ & Mean Roughness $\downarrow$ \\
\midrule
Initial boxes & $0.529{\pm}0.016$ & $1.970{\pm}0.089$ \\
Point-to-plane ICP & $0.860{\pm}0.007$ & $0.200{\pm}0.055$ \\
Frame-wise (no temporal term) & $0.954{\pm}0.007$ & $0.444{\pm}0.084$ \\
\midrule
\rowcolor{gray!12} Kalman RTS & $0.887{\pm}0.042$ & $0.022{\pm}0.003$ \\
\rowcolor{gray!12} Savitzky--Golay & $0.914{\pm}0.024$ & $0.094{\pm}0.014$ \\
\rowcolor{gray!12} \textbf{Ours} (3-term, $\varepsilon{=}0.5$) & $0.946{\pm}0.017$ & $0.077{\pm}0.016$ \\
\bottomrule
\end{tabular}
\end{table}

Table~\ref{tab:baselines} reports the comparison on the idealised multi-face generator. The contribution is not a raw-IoU gain over the initialization; frame-wise registration already recovers most of it, but the accuracy--smoothness \emph{operating point}: among temporally consistent methods ($\bar{R}\le0.1$), joint optimization is the most accurate. Three findings. (i)~\emph{Within the budget}, the margin over the smoothers, the genuine decoupled alternative, is large and consistent on paired per-cell differences: ours vs. SavGol $=+0.032$; and ours vs. Kalman $=+0.059$. (ii)~Against frame-wise registration, ours cuts roughness $5.8\times$ ($0.444\!\to\!0.077$) at a paired IoU difference of $-0.008$; accuracy \emph{matched}, smoothness far better. (iii)~ICP and frame-wise sit outside the budget ($0.200$ and $0.444$); ICP additionally underperforms because its box-surface model cannot exploit the unobserved faces of a partial observation.

\subsection{Joint Optimization Results}\label{sec:richness}

The result above is measured where every face of the object is visible; a real LiDAR sees a vehicle from one side. We therefore treat observation richness as an experimental variable, sweeping $c$ across five regimes on two vehicle shapes, three motions and five seeds (150 tracks), with the joint method and the smoother family each at their best in-budget operating point.

\begin{figure}[t]
\centering
\includegraphics[width=\textwidth]{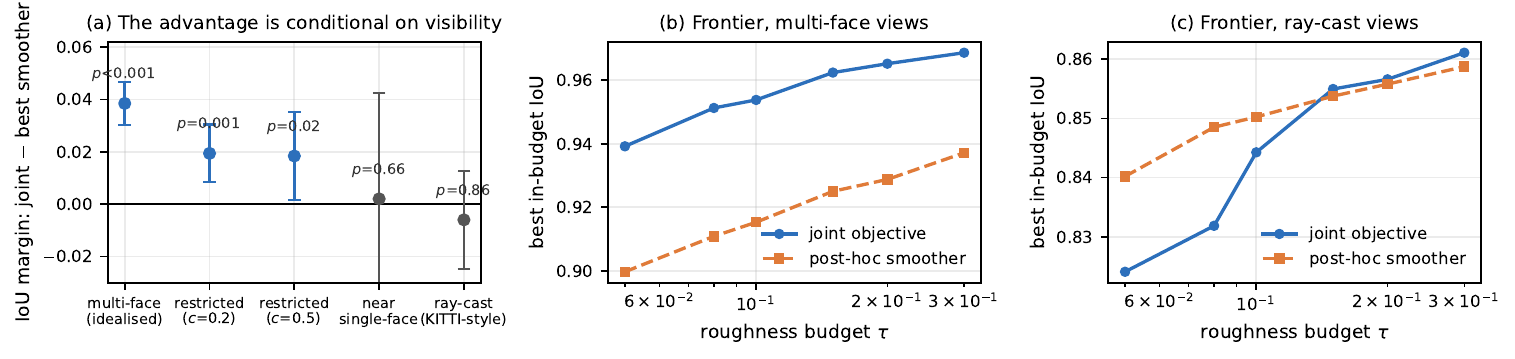}
\caption{The joint-vs-decoupled advantage is conditional on how much of the object is observed. (a)~Paired IoU margin per observation regime with $95\%$ CIs; blue marks $p<0.05$. (b,c)~Accuracy--smoothness frontier in the two extreme regimes ($n{=}30$ tracks each, identical track set at every budget): the joint advantage present under multi-face views is absent, and reversed at tight budgets, under ray-cast views.}\label{fig:regimes}
\end{figure}

Fig.~\ref{fig:regimes}a shows the margin falling from $+0.039$ under multi-face views, through $+0.019$ and $+0.018$ at intermediate visibility, to $+0.002$ when the object is seen almost edge-on, and $-0.006$ on the ray-cast simulator. Across all 150 tracks, the margin correlates with the observed-face count (Spearman $\rho{=}{+}0.31$, $p{=}1.3{\times}10^{-4}$). The advantage established in Sec.~\ref{sec:main} is therefore real but \emph{regime-dependent}, and the idealised generator is its best case.

Fig.~\ref{fig:regimes}b,c trace the frontier --- best IoU within a roughness budget $\tau$ --- in the two extreme regimes. Under multi-face views the joint frontier lies above the decoupled one at every budget ($+0.032$ to $+0.040$). Under ray-cast views the ordering \emph{reverses at tight budgets}: the smoother family leads by $0.016$ at $\tau{=}0.05$, and the curves cross only near $\tau\approx0.15$. Forcing the joint objective into a tight budget under one-sided, range-dependent observations costs more accuracy than smoothing an already-good frame-wise fit. (We exclude $\tau{=}0.02$, where only 20 of 30 tracks admit a feasible solution: including it would compare a strictly easier subset.)

\subsection{Ablation studies}\label{sec:ablation}

\begin{table}[t]
\centering
\caption{Ablation at $\varepsilon{=}0.5$ (SUV, $n{=}15$). $\Delta$ is the paired difference from the full three-term objective.}\label{tab:ablation}
\begin{tabular}{lccc}
\toprule
Configuration & IoU (MC) & $\Delta$ & Roughness $\downarrow$ \\
\midrule
Full (3-term) & 0.946 & --- & 0.077 \\
w/o $\mathcal{L}_c$ (closeness) & 0.938 & $-0.008$  & 0.033 \\
w/o $\mathcal{L}_e$ (enclosure) & \textbf{0.821} & $\mathbf{-0.125}$  & 0.084 \\
w/o $\mathcal{L}_s$ (smoothness) & 0.954 & $+0.008$  & \textbf{0.444} \\
$+\,\mathcal{L}_a$ (4-term) & 0.948 & $+0.002$  & 0.063 \\
\bottomrule
\end{tabular}
\end{table}

Table~\ref{tab:ablation} removes one term at a time at the recommended operating point and reports both metrics, because they disagree. The enclosure term is the accuracy anchor: removing it costs $0.125$ IoU. The smoothness term is the roughness driver: removing it inflates roughness $5.8\times$, confirming it regularises rather than fits. The closeness term is important to both accuracy and smoothness; since the ablation holds $\varepsilon$ fixed, it may overstate a term's necessity: an objective missing one term can compensate by re-tuning the temporal scale. We therefore compared whole objectives, each at \emph{its own} best in-budget operating point, over both observation regimes and two initialization-noise levels ($n{=}18$ tracks per condition). Two-term $\{\mathcal{L}_e,\mathcal{L}_s\}$, three-term and four-term reach respectively $0.947/0.949/0.947$ and $0.953/0.943/0.923$ (idealised, $\times1$ and $\times3$ noise), and $0.837/0.843/0.792$ and $0.830/0.835/0.740$ (ray-cast). The three-term default is nominally best in three of four conditions and never significantly worse.

A single global $\varepsilon$ is suboptimal: some tracks are already smooth under frame-wise registration. Our \emph{ground-truth-free} rule registers frame-wise, measures its roughness $\bar{R}_{\mathrm{fw}}$, keeps frame-wise if $\bar{R}_{\mathrm{fw}}\le0.1$, and otherwise raises $\varepsilon$ to the smallest value entering the budget. Given Sec.~\ref{sec:richness}, we also tested whether the same cues could select \emph{which terms} to include: one variant gates $\delta$ on $\bar{R}_{\mathrm{fw}}$ (threshold fitted once on that sweep), another gates $\theta$ on the observed-face count. Table~\ref{tab:selection} evaluates them on 72 held-out tracks: seeds disjoint from the fitting sweep, and two object classes absent from it. The rule's dependable benefit is \emph{budget compliance}: without ground truth or class labels it keeps every track inside the budget, where a fixed $\varepsilon$ leaves $6\%$ of idealised and $17\%$ of ray-cast tracks outside. On accuracy it is the best non-oracle candidate in the idealised regime.

\begin{table}[t]
\centering
\caption{Operating-point selectors on held-out tracks (4 shapes $\times$ 3 motions $\times$ 3 seeds, both regimes). Selecting the temporal \emph{scale} works; extending selection to \emph{which terms} to use does not.}\label{tab:selection}
\begin{tabular}{lcccc}
\toprule
 & \multicolumn{2}{c}{Idealised} & \multicolumn{2}{c}{Ray-cast} \\
\cmidrule(lr){2-3}\cmidrule(lr){4-5}
Selector & IoU & in-budget & IoU & in-budget \\
\midrule
Best post-hoc smoother & 0.941 & 100\% & \textbf{0.815} & 100\% \\
Fixed $\varepsilon{=}0.5$ (3-term) & 0.949 & 94\% & 0.804 & 83\% \\
\textbf{Adaptive $\varepsilon$ (ours)} & \textbf{0.963} & \textbf{100\%} & 0.801 & \textbf{100\%} \\
\quad + $\delta$ gated on $\bar{R}_{\mathrm{fw}}$ & 0.955 & 100\% & 0.807 & 100\% \\
\quad + $\theta$ gated on face count & 0.948 & 100\% & 0.773 & 100\% \\
\midrule
Oracle (full search) & 0.975 & 100\% & 0.825 & 100\% \\
\bottomrule
\end{tabular}
\end{table}

Across the four shapes and three motions at fixed $\varepsilon{=}0.5$, recovery from the poor initialization is largest for the smallest objects (pedestrian $+0.67$, cyclist $+0.51$, sedan $+0.34$, SUV $+0.31$ IoU); per-cell winners in the highway and urban regimes differ by $0.001$--$0.006$ IoU and are noise-dominated, so the generality argument rests on the pooled statistics and the frontier. IoU degrades $\approx4\%$ from clean to $\sigma{=}0.3$\,m point noise and is stable to $80\%$ occlusion. L-BFGS converges in $\approx35$ iterations ($\approx10$\,s per 30-frame track on one CPU core), with no training and no GPU.

\section{Discussion and Conclusion}\label{sec:conclusion}

We presented a training-free differentiable framework for registering sequential bounding boxes with point cloud streams, geometric constraints (closeness, enclosure) plus a temporal smoothness constraint in one L-BFGS objective, and asked when folding the temporal constraint \emph{inside} the objective beats the decoupled register-then-smooth pipeline.

The answer is conditional, and the condition is observability. Where the sensor sees most of the object, joint optimization is clearly preferable: within the low-roughness budget it beats both post-hoc smoothers ($+0.032$ and $+0.059$ IoU, ahead on every track) and matches frame-wise accuracy at $5.8\times$ lower roughness. As views become one-sided the margin decays, and under a ray-cast simulator it disappears, with the decoupled pipeline ahead at tight budgets. Locating this boundary is the more useful result: it says when the extra coupling is worth its cost, and it explains a contradiction between two of our own simulators that a single-setting evaluation would have hidden. Supporting this, the terms divide cleanly, enclosure anchors accuracy ($-0.125$ IoU when removed), smoothness drives roughness ($5.8\times$, at no accuracy cost), while orientation alignment is fragile for an identifiable reason: it couples yaw to the \emph{estimated} velocity, costing up to $0.09$ IoU when range noise and ego motion corrupt it. A ground-truth-free rule reliably selects the temporal scale; extending the same cues to select which \emph{terms} to include did not work, which we report rather than omit.

\textbf{Limitations.} Evaluation is synthetic-controlled: our ray-cast simulator adds one-sided visibility, self-occlusion, range-dependent density and ego motion, but real LiDAR further contributes sensor-specific noise, clutter and imperfect detections, so no result here is a real-world claim. Initialization is ground-truth perturbation, a fair model of a single-frame detector without tracking, but not of a tracker that already smooths. The framework assumes known, constant box dimensions; sequence endpoints are weakly constrained because the smoothness term needs neighbours; and ultra-tight roughness budgets are unreachable. Our observed-face estimator is monotone but biased in the sparse regime ($2.8$ reported where the truth is $1.2$), supporting a trend rather than a calibrated threshold.

\textbf{Future work.} The decisive experiment is a real-LiDAR study on KITTI tracking; our ray-cast results make its outcome genuinely uncertain, which is why it is worth running. An observation-quality-aware weighting that succeeds where our cue-based gating failed, and handling of unknown box dimensions, would broaden applicability.

\bibliographystyle{splncs04}
\bibliography{references}

@article{besl1992icp,
  title={A method for registration of 3-{D} shapes},
  author={Besl, Paul J and McKay, Neil D},
  journal={IEEE Transactions on Pattern Analysis and Machine Intelligence},
  volume={14},
  number={2},
  pages={239--256},
  year={1992}
}

@article{10056417,
  author={Li, Xuesong and Guivant, Jose E.},
  journal={IEEE Transactions on Intelligent Transportation Systems}, 
  title={Efficient and Accurate Object Detection With Simultaneous Classification and Tracking Under Limited Computing Power}, 
  year={2023},
  volume={24},
  number={6},
  pages={5740-5751},
  doi={10.1109/TITS.2023.3248083}}

@article{yang2016goicp,
  title={Go-{ICP}: A globally optimal solution to 3{D} {ICP} point-set registration},
  author={Yang, Jiaolong and Li, Hongdong and Campbell, Dylan and Jia, Yunde},
  journal={IEEE Transactions on Pattern Analysis and Machine Intelligence},
  volume={38},
  number={11},
  pages={2241--2254},
  year={2016}
}

@article{li2019three,
  title={Three-dimensional backbone network for 3d object detection in traffic scenes},
  author={Li, Xuesong and Guivant, Jose and Kwok, Ngaiming and Xu, Yongzhi and Li, Ruowei and Wu, Hongkun},
  journal={arXiv preprint arXiv:1901.08373},
  year={2019}
}

@article{li2020real,
  title={Real-time 3D object proposal generation and classification using limited processing resources},
  author={Li, Xuesong and Guivant, Jose and Khan, Subhan},
  journal={Robotics and Autonomous Systems},
  volume={130},
  pages={103557},
  year={2020},
  publisher={Elsevier}
}

@inproceedings{qi2021offboard,
  title={Offboard 3{D} object detection from point cloud sequences},
  author={Qi, Charles R and Zhou, Yin and Najibi, Mahyar and Sun, Pei and Vo, Khoa and Deng, Boyang and Anguelov, Dragomir},
  booktitle={CVPR},
  pages={6134--6144},
  year={2021}
}

@article{yang2021auto4d,
  title={Auto4{D}: Learning to label 4{D} objects from sequential point clouds},
  author={Yang, Benjamin and Luo, Wenjie and Urtasun, Raquel},
  journal={arXiv preprint arXiv:2101.06586},
  year={2021}
}

@article{hong2026pointcam,
  title={Pointcam: Cut-and-mix for open-set point cloud learning},
  author={Hong, Jie and Qiu, Shi and Li, Weihao and Anwar, Saeed and Harandi, Mehrtash and Barnes, Nick and Petersson, Lars},
  journal={Computer Vision and Image Understanding},
  pages={104814},
  year={2026},
  publisher={Elsevier}
}

@inproceedings{fan2023ctrl,
  title={Once detected, never lost: Surpassing human performance in offline {LiDAR} based 3{D} object detection},
  author={Fan, Lue and Yang, Yuxue and Wang, Feng and Wang, Naiyan and Zhang, Zhaoxiang},
  booktitle={ICCV},
  pages={3586--3595},
  year={2023}
}

@article{fusion4dal2025,
  title={Fusion4{DAL}: Offline multi-modal 3{D} object detection for 4{D} auto-labeling},
  author={Yang, Zhiyuan and Wang, Xuekuan and Zhang, Wei and Tan, Xiao and Lu, Jincheng and Wang, Jingdong and Ding, Errui and Zhao, Cairong},
  journal={International Journal of Computer Vision},
  year={2025}
}

@inproceedings{openbox2025,
  title={Open{B}ox: Annotate any bounding boxes in 3{D}},
  author={Lee, In-Jae and Kim, Mungyeom and Ryu, Kwonyoung and Musacchio, Pierre and Park, Jaesik},
  booktitle={NeurIPS},
  year={2025}
}

@inproceedings{zopp2024,
  title={{ZOPP}: A framework of zero-shot offboard panoptic perception for autonomous driving},
  author={Ma, Tao and Zhou, Hongbin and Huang, Qiusheng and Yang, Xuemeng and Guo, Jianfei and Zhang, Bo and Dou, Min and Qiao, Yu and Shi, Botian and Li, Hongsheng},
  booktitle={NeurIPS},
  year={2024}
}

@inproceedings{wang2019dcp,
  title={Deep closest point: Learning representations for point cloud registration},
  author={Wang, Yue and Solomon, Justin M},
  booktitle={ICCV},
  pages={3523--3532},
  year={2019}
}

@inproceedings{vedder2025eulerflow,
  title={Neural {E}ulerian scene flow fields},
  author={Vedder, Kyle and Peri, Neehar and Khatri, Ishan and Li, Siyi and Eaton, Eric and Kocamaz, Mehmet and Wang, Yue and Yu, Zhiding and Ramanan, Deva and Pehserl, Joachim},
  booktitle={ICLR},
  year={2025}
}

@inproceedings{li2019detection,
  title={Detection of Imaged Objects with Estimated Scales.},
  author={Li, Xuesong and Kwok, Ngaiming and Guivant, Jose E and Narula, Karan and Li, Ruowei and Wu, Hongkun},
  booktitle={VISIGRAPP (5: VISAPP)},
  pages={39--47},
  year={2019}
}

@inproceedings{diff2i2p2025,
  title={Diff$^2${I2P}: Differentiable image-to-point cloud registration with diffusion prior},
  author={Mu, Juncheng and Ren, Chengwei and Zhang, Weixiang and Pan, Liang and Zhang, Xiao-Ping and Gao, Yue},
  booktitle={ICCV},
  year={2025}
}

@inproceedings{lin2024icpflow,
  title={{ICP-Flow}: {LiDAR} scene flow estimation with {ICP}},
  author={Lin, Yancong and Caesar, Holger},
  booktitle={CVPR},
  pages={14957--14966},
  year={2024}
}

@article{dcreg2026,
  title={{DC-Reg}: Globally optimal point cloud registration via tight bounding with difference of convex programming},
  author={Lian, Wei and Ma, Fei and Pan, Hang and Cui, Zhesen and Zuo, Wangmeng},
  journal={arXiv preprint arXiv:2603.25442},
  year={2026}
}

@inproceedings{rusinkiewicz2001icp,
  title={Efficient variants of the {ICP} algorithm},
  author={Rusinkiewicz, Szymon and Levoy, Marc},
  booktitle={3DIM},
  pages={145--152},
  year={2001}
}

@inproceedings{segal2009gicp,
  title={Generalized-{ICP}},
  author={Segal, Aleksandr and Haehnel, Dirk and Thrun, Sebastian},
  booktitle={RSS},
  year={2009}
}

@book{nocedal2006optimization,
  title={Numerical Optimization},
  author={Nocedal, Jorge and Wright, Stephen J},
  edition={2},
  publisher={Springer},
  year={2006}
}

@article{sun2020waymo,
  title={Scalability in perception for autonomous driving: {Waymo Open Dataset}},
  author={Sun, Pei and Kretzschmar, Henrik and Dotiwalla, Xerxes and Chouard, Aurelien and Patnaik, Vijaysai and Tsui, Paul and Guo, James and Zhou, Yin and Chai, Yuning and Caine, Benjamin and others},
  journal={CVPR},
  pages={2446--2454},
  year={2020}
}

@inproceedings{caesar2020nuscenes,
  title={nu{S}cenes: A multimodal dataset for autonomous driving},
  author={Caesar, Holger and Bankiti, Varun and Lang, Alex H and Vora, Sourabh and Liong, Venice Erin and Xu, Qiang and Krishnan, Anush and Pan, Yu and Baldan, Giancarlo and Beijbom, Oscar},
  booktitle={CVPR},
  pages={11621--11631},
  year={2020}
}

\end{document}